\documentclass[9pt,conference]{IEEEtran}
\IEEEoverridecommandlockouts

\usepackage{cite}
\usepackage{amsmath,amssymb,amsfonts}
\usepackage{algorithmic}
\usepackage{graphicx}
\usepackage{textcomp}
\usepackage{stfloats}
\usepackage{pifont}

\usepackage[table]{xcolor}
\usepackage{booktabs}
\usepackage{bbding}
\usepackage{utfsym}

\begin{document}
\title{PromptTA: Prompt-driven Text Adapter for Source-free Domain Generalization
\thanks{
$^\star$Equal contribution. $^{\dag}$Corresponding author.
This work was supported by the National Natural Science Foundation of China with grant numbers (U21A20485, 62436005, and 62088102), and the Fundamental Research Funds for Xi’an Jiaotong University under Grants xzy022024012.
}
}



\author{
    \IEEEauthorblockN{Haoran Zhang$^{1\star}$, Shuanghao Bai$^{1\star}$, Wanqi Zhou$^{1}$, Jingwen Fu$^{1}$, Badong Chen$^{1\dag}$}
    \IEEEauthorblockA{$^1$Institute of Artificial Intelligence and Robotics, Xi’an Jiaotong University, Xi’an, China}
    \IEEEauthorblockA{\{zhr2001, baishuanghao, zwq785915792, fu1371252069\}@stu.xjtu.edu.cn, chenbd@mail.xjtu.edu.cn}
}

\maketitle

\begin{abstract}
Source-free domain generalization (SFDG) tackles the challenge of adapting models to unseen target domains without access to source domain data. 
To deal with this challenging task, recent advances in SFDG have primarily focused on leveraging the text modality of vision-language models such as CLIP. 
These methods involve developing a transferable linear classifier based on diverse style features extracted from the text and learned prompts or deriving domain-unified text representations from domain banks. 
However, both style features and domain banks have limitations in capturing comprehensive domain knowledge.
In this work, we propose Prompt-Driven Text Adapter (PromptTA) method, which is designed to better capture the distribution of style features and employ resampling to ensure thorough coverage of domain knowledge. 
To further leverage this rich domain information, we introduce a text adapter that learns from these style features for efficient domain information storage.
Extensive experiments conducted on four benchmark datasets demonstrate that PromptTA achieves state-of-the-art performance.
The code is available at https://github.com/zhanghr2001/PromptTA.
\end{abstract}


\section{Introduction}
\begin{figure}[thbp]
\centering
\includegraphics[width=0.473\textwidth]{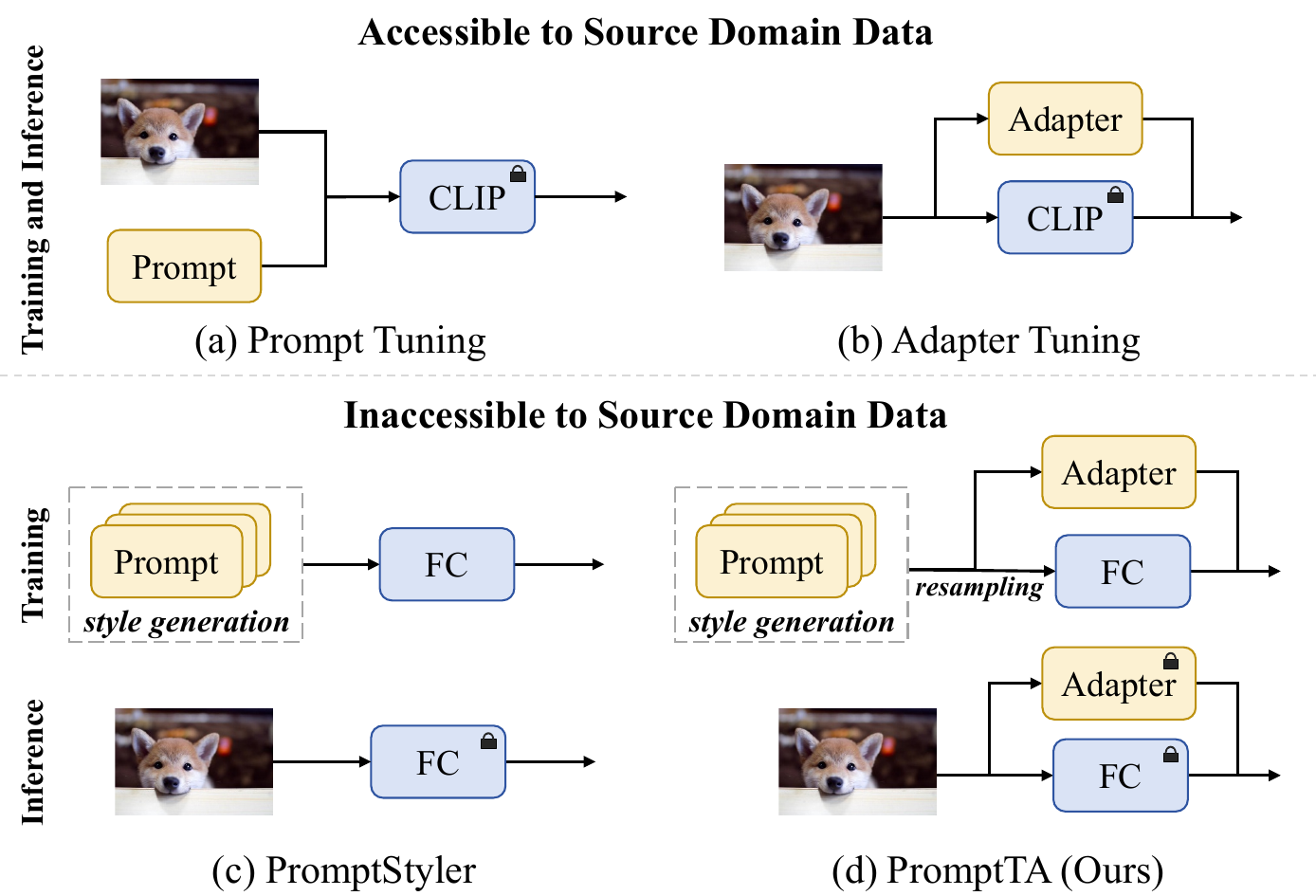}
\caption{Comparison of CLIP-based domain generalization methods.
(a) and (b) require source domain data for fine-tuning, while (c)~\cite{cho2023promptstyler} and our method operate without such data. 
Our method uniquely leverages diverse domain information through style feature resampling and a text adapter.
}
\label{fig:intro}
\end{figure}

Modern deep neural networks often rely on the oversimplified assumption that training and testing data are independent and identically distributed (i.i.d.), which makes them susceptible to out-of-distribution (OOD) data. 
To address the challenge of distribution shift, domain generalization (DG) has been introduced~\cite{gulrajani2020search, cha2022domain, huang2023sentence, bai2024soft}. 
DG aims to develop models based on one or several related but distinct source domains which can generalize well to unseen target domains~\cite{zhou2022domain}.
The emergence of vision-language models (VLMs), like CLIP~\cite{radford2021learning}, has catalyzed the development of promising DG methods, notably those employing prompt tuning~\cite{zhou2022learning, zhou2022conditional, khattak2023maple} and adapter tuning~\cite{zhang2022tip, song2023meta, wang2024caps}, as illustrated in Fig.~\ref{fig:intro} (a) and (b), respectively.
However, DG assumes access to source domain data, which may not be feasible in scenarios involving confidentiality, privacy concerns, or data transmission limitations.

Recently, source-free domain generalization (SFDG) has been proposed to enable the model to make predictions on unseen target domains without requiring source domain data for training. 
Existing SFDG methods primarily focus on leveraging prompt engineering and prompt tuning within VLMs~\cite{niu2022domain, cho2023promptstyler}.
These methods enhance generalization capabilities by utilizing augmented target task definitions (e.g., class names) rather than relying on image data. 
For instance, Niu et al.~\cite{niu2022domain} introduced the concept of domain bank to incorporate textual domain knowledge into soft prompts. 
As shown in Fig.~\ref{fig:intro} (c), Cho et al.~\cite{cho2023promptstyler} proposed PromptStyler, which learns a transferable linear classifier in joint vision-language space by integrating textual style features that encompass diverse domain knowledge.
However, both the textual domain bank and style features have limitations in fully capturing comprehensive domain knowledge, due to constraints in the number of domains represented in the domain bank and the quantity of style features.

In this paper, we propose a method that incorporates a prompt-driven text adapter for source-free domain generalization, namely PromptTA. 
Building upon the style features design of PromptStyler~\cite{cho2023promptstyler}, we establish distributions for these style features and implement resampling to ensure the representation of highly diverse domain knowledge. 
To further leverage this rich domain information, we introduce a text-based adapter that learns from these style features for efficient
domain information storage. The text adapter is initialized with the template ``a [DOM] of a [CLS]", where [DOM] represents the domain name and [CLS] represents the class name. 
Both the resampled and original style features are utilized in training the text adapter and the linear classifier. 
This approach addresses the previous limitations of fully capturing comprehensive domain knowledge and effectively utilizes available domain information.

Our contributions are summarized as follows:

\begin{itemize}
\item We propose PromptTA, a novel adapter-based framework for SFDG that incorporates a text adapter to effectively leverage rich domain information.
\item We introduce style feature resampling that ensures comprehensive coverage of textual domain knowledge.
\item Extensive experiments demonstrate that our PromptTA achieves the state of the art on DG benchmarks.
\end{itemize}

\begin{figure*}[thbp]
\centering
\includegraphics[width=\textwidth]{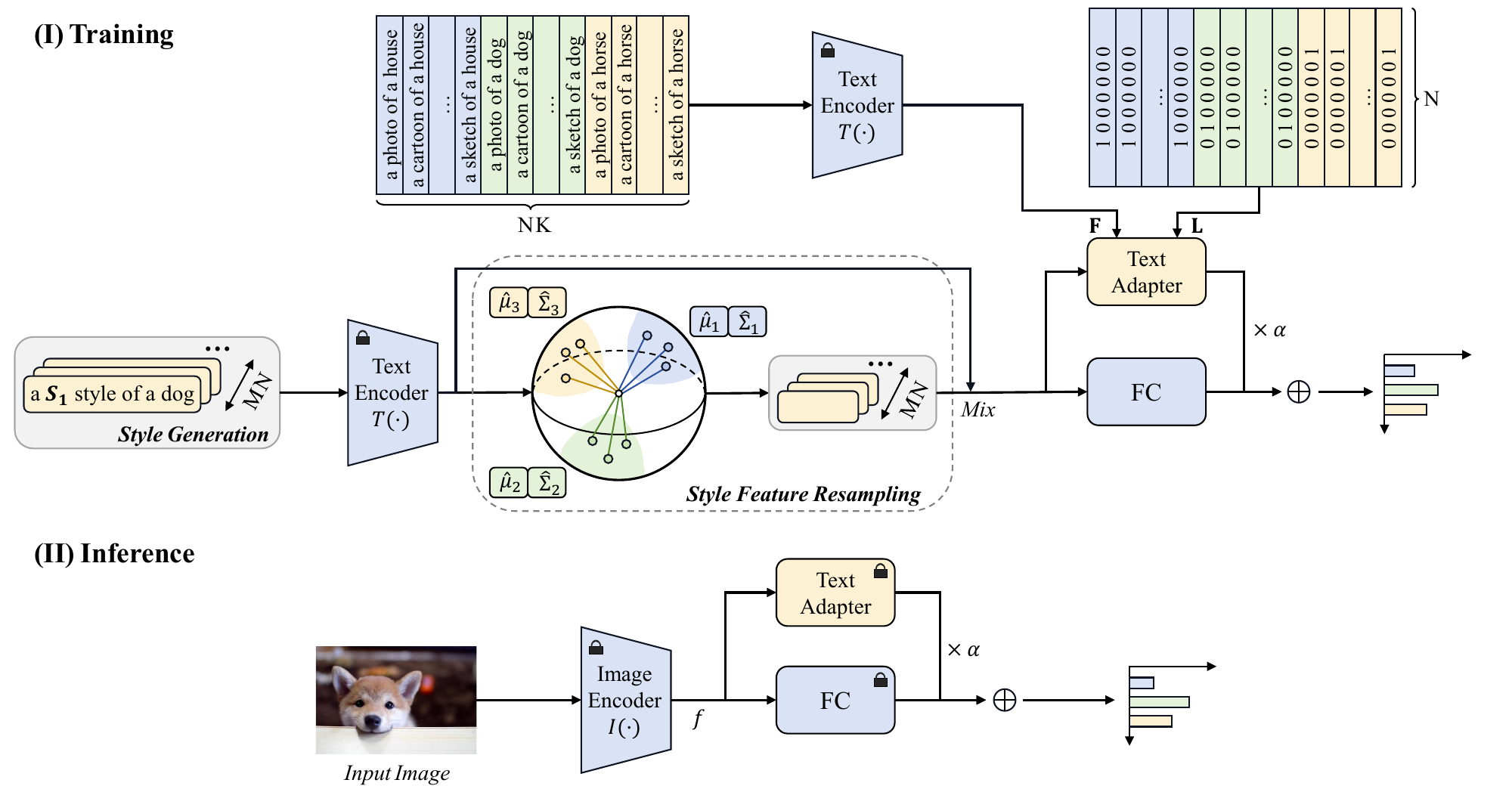}
\caption{Overall framework of PromptTA. Initially, the style generation process yields a fixed set of style features.
These features are then enhanced through Style Feature Resampling to capture comprehensive domain knowledge. 
Both the original style features and resampled style features are utilized to train a linear classifier and a text adapter. 
Note that the encoders are derived from CLIP model~\cite{radford2021learning}.}
\label{fig:model}
\end{figure*}

\section{Related Work}

\noindent \textbf{Domain Generalization.} 
DG aims to train a model on source domain data that effectively generalizes to unseen target domains. 
Most DG methods address distribution shifts through three main perspectives~\cite{zhou2022domain}.
Data augmentation enriches training datasets to help models learn more generalizable features~\cite{volpi2019addressing, shi2020towards}. 
Representation learning designs feature extractors and regularizers to learn domain-invariant features~\cite{li2018deep, arjovsky2019invariant}.
Classifier debiasing retrains classifiers on balanced datasets to enhance generalization~\cite{rosenfeld2022domain, iwasawa2021test}. 
Our method leverages the vision-language alignment properties of CLIP~\cite{radford2021learning}, conceptually aligning more closely with the third perspective by learning a robust classifier in the joint vision-language space.

\noindent \textbf{Adaptation of Vision language Models.} 
Large-scale VLMs~\cite{radford2021learning} have shown impressive zero-shot capabilities, but their size renders full fine-tuning impractical.
Parameter-efficient methods like prompt tuning and adapter tuning have emerged to address this issue. 
Prompt tuning methods, such as CoOp~\cite{zhou2022learning} and CoCoOp~\cite{zhou2022conditional}, replace fixed prompt contexts with learnable vectors, significantly outperforming hand-crafted prompts. 
Adapter tuning methods introduce a small number of additional parameters via residual connections. 
For instance, CLIP-adapter~\cite{gao2024clip} employs residual feature blending with original CLIP-encoded features, while Tip-adapter~\cite{zhang2022tip} utilizes a learnable key-value cache model to refine predictions.
Our method differs from previous works that used adapters composed of visual features. Instead, we address the SFDG task by employing an adapter constructed from text features. 
We further fine-tune the text adapter with a diverse set of learned prompts to enhance the adaptability. 

\noindent \textbf{Source-free Domain Generalization.}
Based on the setting of DG, SFDG tackles a more challenging scenario where source domain data is unavailable.
Current SFDG methods predominantly leverage the alignment between textual and visual features in VLMs, which enhance visual recognition by utilizing rich domain information to learn generalizable text representations. 
Niu et al.~\cite{niu2022domain} employ a pre-defined domain bank to obtain domain-unified text representations, while PromptStyler~\cite{cho2023promptstyler} learns style features covering diverse domains to train a robust linear classifier, which can be transferred for image classification. 
However, these methods are constrained by the limited number of domains in the domain bank or the quantity of style features. 
To address these limitations, we propose style feature resampling to capture comprehensive domain knowledge, coupled with a text adapter for efficient domain information storage.

\begin{table*}[htbp]
\caption{Comparison with the state-of-the-art (SOTA) methods on four domain generalization benchmark datasets for multi-source DG.\\
$\dag$ represents our reproduced results for PromptStyler, which trains the linear classifier
with cross-entropy loss.
}
\centering
\begin{tabular}{
>{\raggedright\arraybackslash}m{3cm}|
>{\centering\arraybackslash}m{2cm}|
>{\centering\arraybackslash}m{1.5cm}|
>{\centering\arraybackslash}m{1.5cm}
>{\centering\arraybackslash}m{1.5cm}
>{\centering\arraybackslash}m{1.5cm}
>{\centering\arraybackslash}m{1.5cm}
>{\centering\arraybackslash}m{1.5cm}
}
\toprule
Method & Venue & Source-free & PACS & VLCS & OfficeHome & DomainNet & Avg. \\
\midrule

\multicolumn{8}{l}{\textit{ResNet-50~\cite{he2016deep} with pre-trained weights on ImageNet~\cite{deng2009imagenet}.}} \\
\midrule
RSC~\cite{huang2020self} & ECCV 2020 & \ding{55} & 85.2\scriptsize{$\pm{0.9}$} & 77.1\scriptsize{$\pm{0.5}$} & 65.5\scriptsize{$\pm{0.9}$} & 38.9\scriptsize{$\pm{0.5}$} & 66.7 \\
SagNet~\cite{nam2021reducing} & CVPR 2021 & \ding{55} & 86.3\scriptsize{$\pm{0.2}$} & 77.8\scriptsize{$\pm{0.5}$} & 68.1\scriptsize{$\pm{0.1}$} & 40.3\scriptsize{$\pm{0.1}$} & 68.1 \\
MIRO~\cite{cha2022domain} & ECCV 2022 & \ding{55} & 85.4\scriptsize{$\pm{0.4}$} & 79.0\scriptsize{$\pm{0.0}$} & 70.5\scriptsize{$\pm{0.4}$} & 44.3\scriptsize{$\pm{0.2}$} & 69.8 \\
SWAD~\cite{cha2021swad} & NeurIPS 2021 & \ding{55} & \textbf{88.1}\scriptsize{$\pm{0.1}$} & \textbf{79.1}\scriptsize{$\pm{0.1}$} & \textbf{70.6}\scriptsize{$\pm{0.2}$} & \textbf{46.5}\scriptsize{$\pm{0.1}$} & \textbf{71.1} \\
\midrule

\multicolumn{8}{l}{\textit{ResNet-50~\cite{he2016deep} with pre-trained weights from CLIP~\cite{radford2021learning}.}} \\
\midrule
ZS-CLIP (C)~\cite{radford2021learning} & ICML 2021 & \ding{51} & 91.0\scriptsize{$\pm{0.0}$} & 81.2\scriptsize{$\pm{0.0}$} & 67.1\scriptsize{$\pm{0.0}$} & 45.9\scriptsize{$\pm{0.0}$} & 71.3 \\
CAD~\cite{ruan2022optimal} & ICLR 2022 & \ding{55} & 90.0\scriptsize{$\pm{0.6}$} & 81.2\scriptsize{$\pm{0.6}$} & 70.5\scriptsize{$\pm{0.3}$} & 45.5\scriptsize{$\pm{2.1}$} & 71.8 \\
ZS-CLIP (PC)~\cite{radford2021learning} & ICML 2021 & \ding{51} & 90.8\scriptsize{$\pm{0.0}$} & 81.4\scriptsize{$\pm{0.0}$} & 70.8\scriptsize{$\pm{0.0}$} & 46.7\scriptsize{$\pm{0.0}$} & 72.4 \\
PromptStyler$^\dag$~\cite{cho2023promptstyler} & ICCV 2023 & \ding{51} & 92.5\scriptsize{$\pm{0.1}$} & 82.2\scriptsize{$\pm{0.1}$} & 72.3\scriptsize{$\pm{0.1}$} & 48.6\scriptsize{$\pm{0.0}$} & 73.9 \\
\cellcolor{gray!9.0}{PromptTA (Ours)} & \cellcolor{gray!9.0}\textbf{--} & \cellcolor{gray!9.0}{\ding{51}} & \cellcolor{gray!9.0}\textbf{93.8}\scriptsize{$\pm{0.0}$} & \cellcolor{gray!9.0}\textbf{83.2}\scriptsize{$\pm{0.1}$} & \cellcolor{gray!9.0}\textbf{73.2}\scriptsize{$\pm{0.1}$} & \cellcolor{gray!9.0}\textbf{49.2}\scriptsize{$\pm{0.0}$} & \cellcolor{gray!9.0}\textbf{74.9} \\
\midrule

\multicolumn{8}{l}{\textit{ViT-B\,/\,16~\cite{dosovitskiy2020vit} with pre-trained weights from CLIP~\cite{radford2021learning}.}} \\
\midrule
ZS-CLIP (C)~\cite{radford2021learning} & ICML 2021 & \ding{51} & 95.8\scriptsize{$\pm{0.0}$} & 76.5\scriptsize{$\pm{0.0}$} & 79.3\scriptsize{$\pm{0.0}$} & 57.3\scriptsize{$\pm{0.0}$} & 77.2 \\
MIRO~\cite{cha2022domain} & ECCV 2022 & \ding{55} & 95.6 & 82.2 & 82.5 & 54.0 & 78.6 \\
ZS-CLIP (PC)~\cite{radford2021learning} & ICML 2021 & \ding{51} & 96.1\scriptsize{$\pm{0.0}$} & 83.4\scriptsize{$\pm{0.0}$} & 81.8\scriptsize{$\pm{0.0}$} & 57.2\scriptsize{$\pm{0.0}$} & 79.6 \\
PromptStyler$^\dag$~\cite{cho2023promptstyler} & ICCV 2023 & \ding{51} & 97.2\scriptsize{$\pm{0.1}$} & 83.4\scriptsize{$\pm{0.3}$} & 82.5\scriptsize{$\pm{0.2}$} & 58.3\scriptsize{$\pm{0.0}$} & 80.4 \\
\cellcolor{gray!9.0}{PromptTA (Ours)} & \cellcolor{gray!9.0}\textbf{--} & \cellcolor{gray!9.0}{\ding{51}} & \cellcolor{gray!9.0}\textbf{97.3}\scriptsize{$\pm{0.1}$} & \cellcolor{gray!9.0}\textbf{83.6}\scriptsize{$\pm{0.3}$} & \cellcolor{gray!9.0}\textbf{82.9}\scriptsize{$\pm{0.0}$} & \cellcolor{gray!9.0}\textbf{59.4}\scriptsize{$\pm{0.0}$} & \cellcolor{gray!9.0}\textbf{80.8} \\
\midrule

\multicolumn{8}{l}{\textit{ViT-L\,/\,14~\cite{dosovitskiy2020vit} with pre-trained weights from CLIP~\cite{radford2021learning}.}} \\
\midrule
ZS-CLIP (C)~\cite{radford2021learning} & ICML 2021 & \ding{51} & 97.7\scriptsize{$\pm{0.0}$} & 79.1\scriptsize{$\pm{0.0}$} & 85.6\scriptsize{$\pm{0.0}$} & 62.8\scriptsize{$\pm{0.0}$} & 81.3 \\
ZS-CLIP (PC)~\cite{radford2021learning} & ICML 2021 & \ding{51} & \textbf{98.6}\scriptsize{$\pm{0.0}$} & 82.6\scriptsize{$\pm{0.0}$} & 86.7\scriptsize{$\pm{0.0}$} & 63.4\scriptsize{$\pm{0.0}$} & 82.8 \\
PromptStyler$^\dag$~\cite{cho2023promptstyler} & ICCV 2023 & \ding{51} & \textbf{98.6}\scriptsize{$\pm{0.0}$} & 82.9\scriptsize{$\pm{0.5}$} & 88.4\scriptsize{$\pm{0.1}$} & 64.5\scriptsize{$\pm{0.0}$} & 83.6 \\
\cellcolor{gray!9.0}{PromptTA (Ours)} & \cellcolor{gray!9.0}\textbf{--} & \cellcolor{gray!9.0}{\ding{51}} & \cellcolor{gray!9.0}\textbf{98.6}\scriptsize{$\pm{0.0}$} & \cellcolor{gray!9.0}\textbf{83.3}\scriptsize{$\pm{0.3}$} & \cellcolor{gray!9.0}\textbf{88.5}\scriptsize{$\pm{0.0}$} & \cellcolor{gray!9.0}\textbf{65.2}\scriptsize{$\pm{0.0}$} & \cellcolor{gray!9.0}\textbf{83.9} \\
\bottomrule
\end{tabular}
\label{table:main_result}
\end{table*}

\section{Method}

The overall framework of the proposed PromptTA method is illustrated in Fig.~\ref{fig:model}. 
Our method utilizes the aligned image-text representations from CLIP~\cite{radford2021learning} to tackle the DG problem without using any images. 
We introduce our method as follows.

\noindent \textbf{Style Generation.}
Following Algorithm 1 of PromptStyler~\cite{cho2023promptstyler}, we employ pseudo-word $S_i$ as a placeholder within the prompt template $P_i$, represented as ``a $S_i$ style of a [CLS]," where [CLS] denotes the class name.
We define $P_i^{dom}$ as ``a $S_i$ style of a" and $P_j^{cls}$ as ``CLS", where $j \in {1, 2, ..., N}$, and $N$ denotes the number of class names.
The pseudo-word $S_i$ is replaced with $M$ learnable style word vectors $\{s_i\}_{i=1}^{M}$ that represent diverse domain information. 
Thus we can derive $MN$ style features corresponding to $\{\{P_{i,j}\}_{i=1}^{M}\}_{j=1}^{N}$.
To optimize these learnable vectors $\{s_i\}_{i=1}^{M}$, a style diversity loss $\mathcal{L}_{\mathrm{style}}$ (Equation 1 in~\cite{cho2023promptstyler}) is adopted to maximize the variance of $M$ text features of $P_i^{dom}$. 
And a content consistency loss $\mathcal{L}_{\mathrm{content}}$ (Equation 3 in~\cite{cho2023promptstyler}) is used to ensure the class information remains consistent by minimizing the distance of text features between $P_{i,j}$ and its corresponding $P_j^{cls}$.
Finally, we obtain a set of fixed learned style word vectors $\{s_i\}_{i=1}^{M}$. The \textit{i}-th style feature of \textit{j}-th class is denoted as $T(P_{i,j})$, where $T(\cdot)$ denotes the text encoder. 

\noindent \textbf{Style Feature Resampling.}
While PromptStyler~\cite{cho2023promptstyler} employs a fixed set of $M$ style features to train a linear classifier, this approach has limitations in fully capturing comprehensive domain knowledge. 
To this end, we propose Style Feature Resampling (SFR), a module that incorporates more diverse domain information by dynamically regenerating sampled features during each training epoch, thereby enhancing generalization capabilities.
Building upon the previous subsection, we leverage $M$ learned style word vectors and $N$ classes to derive a total of $MN$ style features. Specifically, the style feature corresponding to the \textit{i}-th style and \textit{j}-th class is denoted as $T(P_{i,j})$.
We posit that for a given class $j$, the prompt features generated by combining each class with various style word vectors follow a Gaussian distribution, as expressed as follows:


\begin{equation}
    \{T(\mathcal\{{P}_{i,j}\}_{i=1}^M)\}\sim\mathcal{N}(\boldsymbol{\mu}_j,\boldsymbol{\varSigma}_j).
\label{eq:hypo}
\end{equation}

The ground-truth feature distribution of the \textit{j}-th class can be approximated by computing mean and standard deviations from known prompt features as follows:

\begin{equation}
    \hat{\boldsymbol{\mu}}_j=\frac{1}{M}\sum_{i=1}^MT(\mathcal{P}_{i,j}),
\end{equation}
\begin{equation}
    \hat{\boldsymbol{\varSigma}}_j=\frac{1}{M-1}\sum_{i=1}^M(T(\mathcal{P}_{i,j}))-\hat{\boldsymbol{\mu}}_j)^2.
\end{equation}

By drawing new features from this estimated distribution, we incorporate new styles that encompass diverse domains. 
We resample style features at the onset of each training iteration, which are then combined with the original features to train the model.

\noindent \textbf{Text Adapter.}
To fully leverage the acquired domain knowledge, we introduce a learnable text adapter that learns from style features. 
Different from previous works~\cite{zhang2022tip}, our adapter is constructed using text features rather than visual features, a design tailored to address the challenges of the SFDG task.
The adapter stores domain knowledge from learning from style features and generates predictions based on the similarities between input features and adapter features. 
To initialize the $N$-way $K$-shot text adapter, we employ a pre-defined domain bank comprising $K$ distinct domains (e.g., photo, painting, sketch).
Following the template ``a [DOM] of a [CLS]", we construct $KN$ prompts and extract adapter features using CLIP's text encoder, resulting in $\mathbf{F}\in\mathbb{R}^{NK\times{D}}$, where $D$ denotes the adapter feature dimension. The classes are encoded as one-hot vectors, forming $\mathbf{L}\in\mathbb{R}^{NK\times{N}}$.
For an input feature $f\in\mathbb{R}^{D}$, we compute its similarities with adapter as follows:

\begin{equation}
\textrm{Logits}_\text{Adapter}=\varphi(f\mathbf{F}^T)\mathbf{L},
\label{eq:adapter}
\end{equation}
\begin{equation}
\varphi(x)=\exp(-\beta(1-x)),
\label{eq:adapter_hyper}
\end{equation}
where $\varphi(\cdot)$ denotes an exponential function that transforms similarity scores into non-negative values. The parameter $\beta$ serves to modulate the sharpness of this transformation, effectively controlling the sensitivity of the adapter to input similarities.
We retain the linear classifier to operate directly on style features, whose weight can be denoted as $W$.
Thus the final classification logits are computed as a weighted combination of the outputs from both the text adapter and the linear classifier, which can be formulated as:

\begin{equation}\begin{aligned}  
\textrm{Logits}
&=\textrm{Logits}_{\mathrm{FC}} + \alpha\textrm{Logits}_{\mathrm{Adapter}}, \\
&=fW^T+\alpha\varphi(f\mathbf{F}^T)\mathbf{L},
\label{eq:logits_sum}
\end{aligned}\end{equation}
where $\alpha$ denotes the residual ratio.
Then we adopt cross-entropy loss as
our classification loss to train the linear classifier and text adapter.
This approach integrates the classification power of the linear classifier with the potential feature refinement capabilities of the text adapter, enhancing the model's performance across diverse domains.

\noindent \textbf{Inference.} 
During inference, an input image is processed by the CLIP image encoder to produce an image feature, which is then passed through both the trained linear classifier and trained text adapter. 
The final prediction is also formulated as~\eqref{eq:logits_sum}.

\begin{figure}[thbp]
\centering
\includegraphics[width=0.47\textwidth]{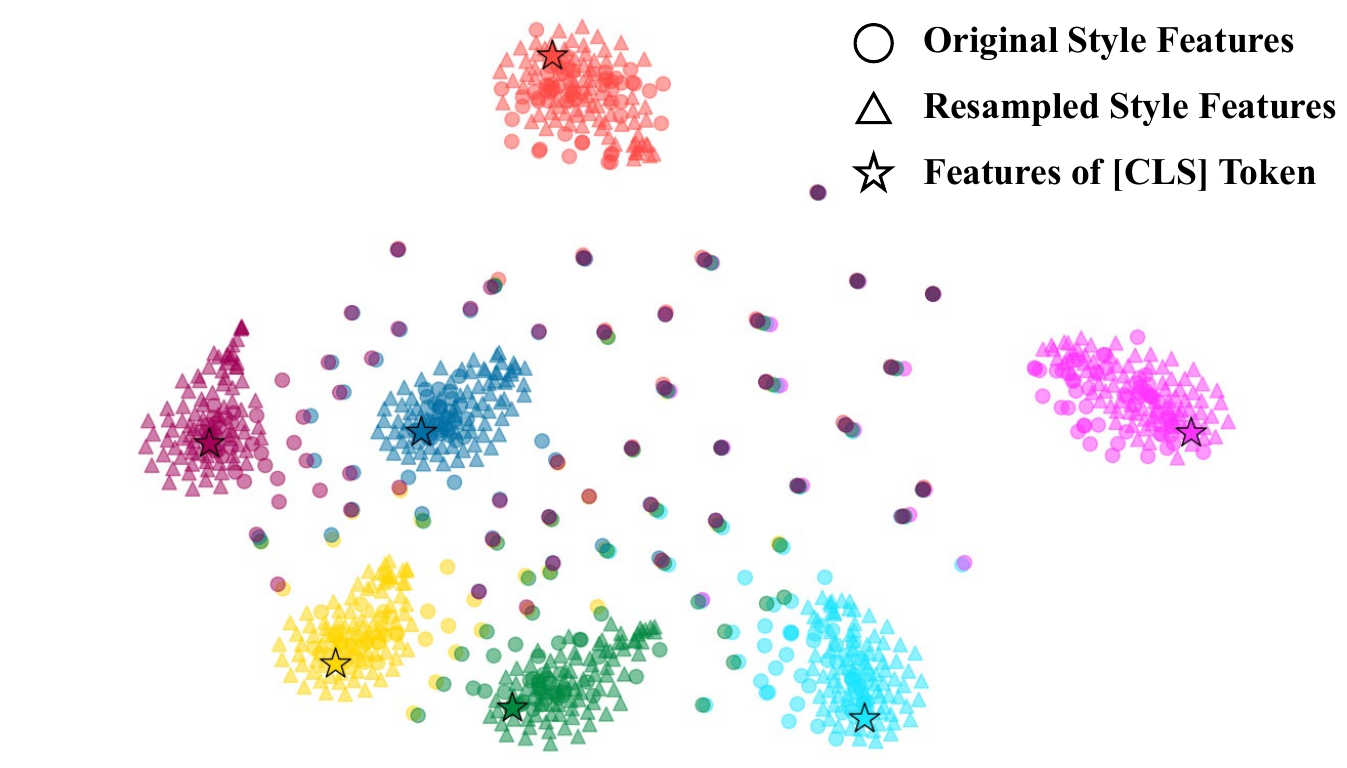}
\caption{The t-SNE~\cite{van2008visualizing} visualization results for original style features, style features from a single resampling instance, and class token features of PACS dataset. Different colors denote different classes.
}
\label{fig:tsne}
\end{figure}

\begin{table}[thbp]
\caption{Ablation study on the model component of style feature resampling~(SFR) and text adapter~(TA).}
\centering
\begin{tabular}{cccccccc}
\toprule

SFR & TA & PACS & VLCS & OfficeHome & DomainNet \\
\midrule
 &  & 92.5 & 82.2 & 72.3 & 48.6 \\
\ding{51} &  & 93.2 & 82.9 & 72.7 & 48.7 \\
 & \ding{51} & 93.6 & 83.1 & 73.1 & 49.0 \\
\cellcolor{gray!9.0}\ding{51} & \cellcolor{gray!9.0}\ding{51} & \cellcolor{gray!9.0}\textbf{93.8} & \cellcolor{gray!9.0}\textbf{83.2} & \cellcolor{gray!9.0}\textbf{73.2} & \cellcolor{gray!9.0}\textbf{49.2} \\
\bottomrule
\end{tabular}
\label{table:ablation}
\end{table}

\section{Experiments}

\subsection{Experimental Settings}

\noindent \textbf{Evaluation datasets.}
To validate the effectiveness of our method, we evaluate it on four DG datasets: PACS~\cite{Li_2017_ICCV}, VLCS~\cite{fang2013unbiased}, OfficeHome~\cite{venkateswara2017deep} and DomainNet~\cite{peng2019moment}. On each dataset, we repeatedly test three times with different random seeds and adopt the average top-1 accuracy with standard deviations.

\noindent \textbf{Baselines.}
We compare our method with a series of baselines, including source-free methods like zero-shot CLIP~\cite{radford2021learning}, PromptStyler~\cite{cho2023promptstyler},  and conventional DG algorithms that require source data: RSC~\cite{huang2020self}, SagNet~\cite{nam2021reducing}, MIRO~\cite{cha2022domain}, SWAD~\cite{cha2021swad}, and CAD~\cite{ruan2022optimal}. 
Note that ZS-CLIP (C) and ZS-CLIP (PC) denote zero-shot CLIP~\cite{radford2021learning} with a prompt template of ``[CLS]" and ``a photo of a [CLS]", respectively.
 
\noindent \textbf{Implementation details.}
During style generation, we adopt the configuration from PromptStyler~\cite{cho2023promptstyler} to learn $M=80$ style word vectors initialized from a zero-mean Gaussian distribution with a standard deviation of 0.02.
The text adapter incorporates $K=11$ domains, including photo, cartoon, and painting, etc.
For training, we employ SGD with 0.9 momentum and a cosine learning rate scheduler. The learning rates are set to 0.05 for the linear classifier and 0.01 for the adapter on PACS, VLCS, and OfficeHome. For DomainNet, we reduce the adapter's learning rate to 0.001 due to its large sample size.
The hyper-parameter $\alpha$ of the text adapter is dataset-specific, ranging from 1 to 5, while $\beta$ is fixed at 2. 
For PromptStyler, we train the linear classifier using cross-entropy loss.

\subsection{Evaluations}

\begin{figure}[ht]
\centering
\includegraphics[width=0.45\textwidth]{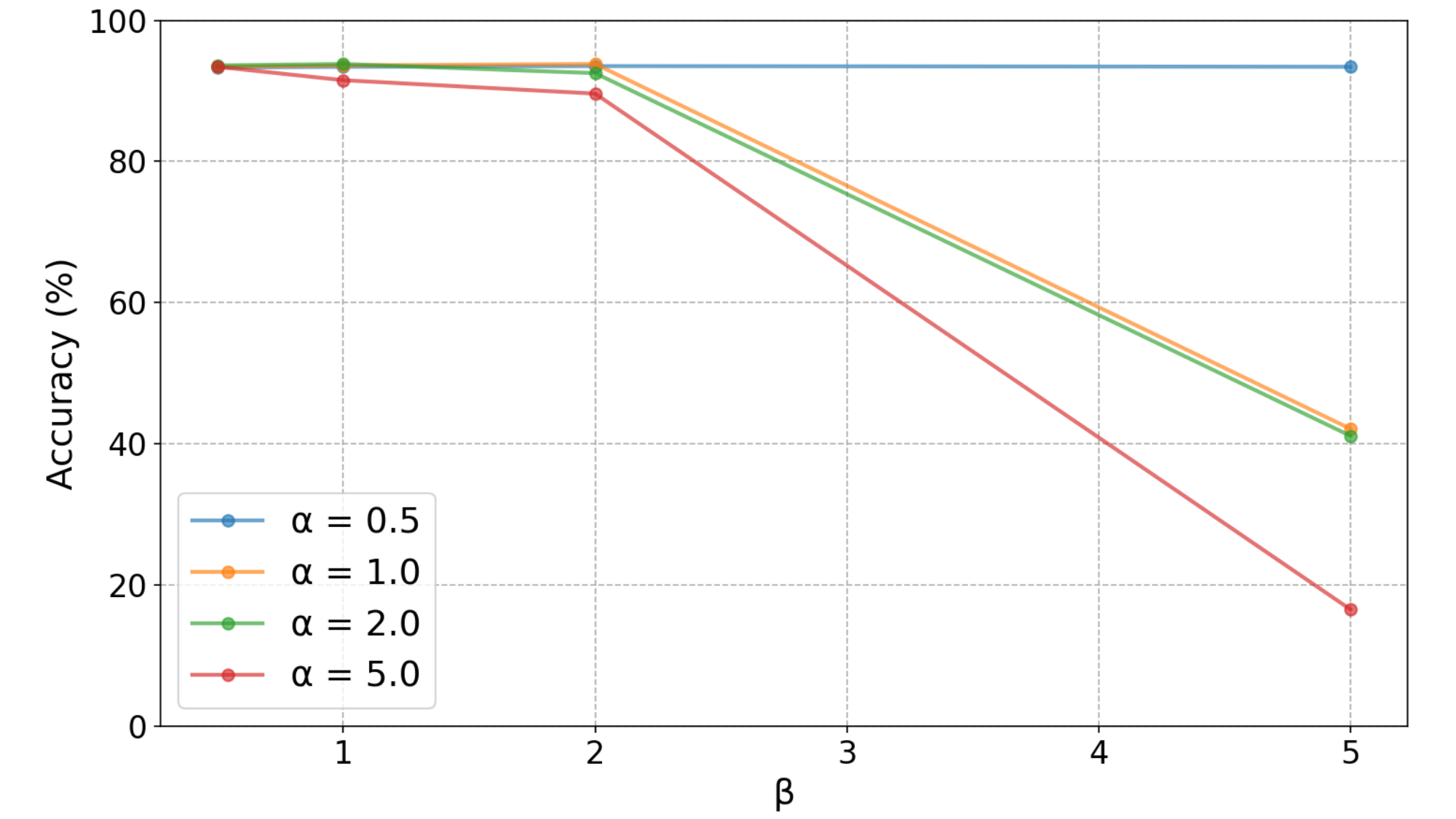}
\caption{Sensitivity analysis of hyperparameter $\alpha$ and $\beta$ on PACS dataset. 
}
\label{fig:sensitivity}
\end{figure}

\begin{table}[ht]
\caption{Comparison of initialization methods for text adapter.}
\centering
\begin{tabular}{ccccccc}
\toprule
Initialization  & PACS & VLCS & OfficeHome & DomainNet \\
\midrule
Random & 93.3 & 82.5 & 72.7 & 47.0 \\
\cellcolor{gray!9.0}Template & \cellcolor{gray!9.0}\textbf{93.8} & \cellcolor{gray!9.0}\textbf{83.2} & \cellcolor{gray!9.0}\textbf{73.2} & \cellcolor{gray!9.0}\textbf{49.2} \\
\bottomrule
\end{tabular}
\label{table:random_initialize}
\end{table}

\noindent \textbf{Main results.}
As shown in Table~\ref{table:main_result}, our PromptTA method achieves SOTA performance across all evaluation datasets with three visual backbones: ResNet-50~\cite{he2016deep}, ViT-B/16~\cite{dosovitskiy2020vit}, and ViT-L/14~\cite{dosovitskiy2020vit}. 
Specifically, PromptTA achieves average accuracy improvements of 1.0\%, 0.4\%, and 0.3\%, respectively, over SOTA PromptStyler method.
Notably, it surpasses conventional DG baselines despite not utilizing any images during training, outperforming SOTA SWAD and CAD methods by 3.8\% and 3.1\% in average accuracy, respectively, with ResNet-50 as the backbone.

\noindent \textbf{T-SNE visualization results.}
As shown in Fig.~\ref{fig:tsne}, for each class, the resampled style features (triangles) align closely with the original style features (circles), indicating that the resampling effectively captures the distribution of the original style features while possibly enhancing generalization. 
Both the original (circles) and resampled (triangles) style features cluster distinctly by class (stars). This suggests that the resampling process maintains the inherent structure of the style features and does not distort the original class boundaries.

\subsection{More analyses}

\noindent \textbf{Ablation study on each module.}
We perform an ablation study on style feature resampling (SFR) and text adapter (TA) to assess their individual and combined contributions.
As shown in Table~\ref{table:ablation}, the removal of either SFR or TA leads to a decrease in accuracy. 
Notably, the highest accuracy is achieved when both components are present, underscoring their synergistic effect on the model's performance.

\noindent \textbf{Sensitivity analysis of hyperparameter $\alpha$ and $\beta$.}
As shown in Fig.~\ref{fig:sensitivity}. We vary $\alpha$ and $\beta$ from 0.5 to 5.0. 
The model exhibits robust performance when $\beta \in [0.5, 2]$ across a wide range of $\alpha$ values $(0.5 \leq \alpha \leq 5)$. For $\beta > 2$, robustness is maintained only when $\alpha \approx 0.5$, while performance degrades significantly for $\alpha \in [1, 5]$.

\noindent \textbf{Ablation study on initialization method of text adapter.}
As shown in Table~\ref{table:random_initialize}, initializing the text adapter with a template (a [DOM] of a [CLS]) as prior knowledge outperforms random initialization. 
This suggests that incorporating domain-specific and class-specific information during initialization provides a better starting point.

\section{Conclusion}
In this paper, we propose PromptTA, a novel method that incorporates text adapter to address the challenging SFDG task. 
We introduce a style feature resampling module to better capture the distribution of style features and employ resampling to ensure comprehensive domain coverage. 
Then a text adapter is utilized to act as a dynamic repository for domain knowledge, which can be effectively leveraged during inference.
Experiments on four benchmarks demonstrate that PromptTA achieves state-of-the-art performance.

\bibliography{IEEEabrv, main}

\end{document}